\documentclass{article}

\usepackage[backend=biber,style=numeric]{biblatex}
\title{MoME: \underline{M}ixture \underline{o}f Visual Language \underline{M}edical \underline{E}xperts for Medical Imaging Segmentation}

\author{
  Arghavan Rezvani$^{1}$ \thanks{Co-first authors.} \\
  \texttt{rezvanid@uci.edu}
  \and
  Xiangyi Yan$^{1}$ \footnotemark[1] \\
  \texttt{xiangyy4@uci.edu}
  \and
  Anthony T. Wu$^{1,2}$ \footnotemark[1]\\
  \texttt{wuat2@hs.uci.edu}
  \and
  Kun Han$^{1}$ \\
  \texttt{khan7@uci.edu}
  \and
  Pooya Khosravi$^{1,2}$ \\
  \texttt{pooyak@hs.uci.edu}
  \and
  Xiaohui Xie$^{1}$ \\
  \texttt{xhx@uci.edu}
}

\date{%
$^{1}$Department of Computer Science, University of California, Irvine\\%
$^{2}$School of Medicine, University of California, Irvine\\
}

\usepackage{geometry}
\usepackage{amsmath, amssymb}
\usepackage{graphicx}
\usepackage{hyperref}
\usepackage{url}

\usepackage{tabularx} 
\usepackage{makecell} 
\usepackage{booktabs} 
\usepackage{multirow}
\usepackage{amsmath,amssymb}

\usepackage{array} 

\newcolumntype{P}[1]{>{\centering\arraybackslash}p{#1}}

\begin{document}

\maketitle
\begin{abstract}
In this study, we propose \textbf{MoME}, a \textbf{M}ixture \textbf{o}f Visual Language \textbf{M}edical \textbf{E}xperts, for Medical Image Segmentation. MoME adapts the successful Mixture of Experts (MoE) paradigm, widely used in Large Language Models (LLMs), for medical vision-language tasks. The architecture enables dynamic expert selection by effectively utilizing multi-scale visual features tailored to the intricacies of medical imagery, enriched with textual embeddings. This work explores a novel integration of vision-language models for this domain. Utilizing an assembly of 10 datasets, encompassing 3,410 CT scans, MoME demonstrates strong performance on a comprehensive medical imaging segmentation benchmark. Our approach explores the integration of foundation models for medical imaging, benefiting from the established efficacy of MoE in boosting model performance by incorporating textual information. Demonstrating competitive precision across multiple datasets, MoME explores a novel architecture for achieving robust results in medical image analysis.

\end{abstract}

\section{Introduction}
\label{sec:intro}
Medical image segmentation (MIS) serves as a vital link between advanced clinical practice and cutting-edge computational methods, enabling a range of critical applications, from early disease detection to precise treatment planning \cite{gibson2018automatic}. While traditional approaches like atlas-based segmentation have advanced MIS, their generalizability and clinical utility remain limited, primarily due to inherent anatomical variability, surgical interventions, and diverse pathological presentations that complicate consistent interpretation.

The advent of deep learning revolutionized MIS, with Convolutional Neural Networks (CNNs) \cite{ronneberger2015unet} driving major progress through their powerful feature extraction capabilities. However, CNNs are inherently limited in modeling long-range dependencies, which are often critical in complex medical imaging tasks. Transformers, originally developed for Natural Language Processing \cite{vaswani2017attention}, address this limitation by effectively capturing long-distance relationships within data, making them well suited to overcome CNN's shortcomings in MIS.

Recently, the development of foundation models for medical imaging has gained momentum. Due to the high cost and effort associated with medical image annotation, researchers have increasingly explored self-supervised learning using large volumes of unlabeled data. 
However, recent evidence continues to demonstrate the superiority of fully supervised pre-training over self-supervised methods when applied at similar scales of CT data \cite{liu2023clipdriven}. This improved efficiency stems from dense supervision and better alignment between the pretext and downstream tasks. In contrast to self-supervised proxy tasks such as jigsaw puzzle solving \cite{noroozi2016unsupervised}, rotation prediction \cite{komodakis2018unsupervised}, or patch location prediction \cite{doersch2015unsupervised}, which often lack strong relevance to segmentation, fully supervised learning directly optimizes for the target task.

In addressing foundation models for MIS, researchers continue to face significant challenges: extreme diversity of anatomical structures, the complex contextual relationships between them, and the need for adaptive processing for different structures with different scales. While current approaches have made progress, we propose a fundamentally different solution through \textbf{MoME, a Mixture of Visual Language Medical Experts for MIS}. Our approach adapts the proven success of Mixture of Experts (MoE) \cite{jiang2024mixtral, xue2024openmoe} to medical imaging by developing a specialized architecture where multiple expert decoders process multi-scale visual features, dynamically weighted by a text-guided routing mechanism. MoME adaptively emphasizes the most relevant visual scales for each structure, guided by textual semantic embeddings. This innovative integration of MoE with visual-language modeling represents a significant advancement in foundation models for MIS.

Amidst the proliferation of public medical image datasets, our work leverages this extensive resource, employing 10 datasets comprising 3,410 CT scans, trained in a fully supervised manner. This comprehensive training approach enhances model performance and significantly contributes to the advancement of MIS technology.


Our contributions are manifold; 
(1) We introduce the first MoE-based vision-language foundation model specifically designed for MIS, demonstrating a novel application of MoE in this domain,
(2) We propose a dynamic expert routing mechanism that effectively integrates multi-scale visual features with textual embeddings, enabling adaptive processing tailored to different anatomical structures,
(3) By training on a diverse collection of public datasets, we ensure that our model is robust and versatile across various segmentation tasks, and
(4) Through conducting extensive experiments, we show that our model achieves state-of-the-art performance across a wide range of public and proprietary datasets, setting new benchmarks in accuracy and efficiency for medical image segmentation. 

This work not only advances the state-of-the-art, but also paves the path for future research at the intersection of medical imaging and machine learning.

\section{Related Work}
\label{sec:relatedwork}

\subsection{Foundational Models in Medical Imaging}
Recent explorations of Foundation models in medical imaging can be broadly divided into two categories: models with only visual prompts and models with additional textual cues.


\textbf{\textit{Visual Prompted Models:}} Clinical-BERT \cite{Yan_Pei_2022} is a vision-language model for radiograph diagnosis and report generation, using domain-specific tasks like Clinical Diagnosis and Image-MeSH Matching, highlighting the value of domain-specific pre-training.
Med-SA \cite{wu2023medical} enhances the Segment Anything Model (SAM) for MIS with domain-specific knowledge, achieving superior performance across 17 tasks with minimal parameter increase. MedSAM \cite{ma2023segment}, the first foundation model for universal MIS, outperforms existing models and matches specialized ones for various tasks. Virchow \cite{vorontsov2024virchow} sets a new standard in pathology image analysis with high accuracy in pan-cancer detection. Swin UNETR \cite{tang2022self}, a self-supervised learning framework for 3D medical image analysis, demonstrates state-of-the-art performance by leveraging self-supervised pre-training on diverse CT image datasets.

\textbf{\textit{Textual Prompted Models:}}
Text data offers rich contextual information that can guide visual representation learning. CheXzero \cite{ExpertlevelTiu} leverages self-supervised learning for chest X-ray interpretation without manual annotations, achieving classification accuracies comparable to those of radiologists. MedCLIP \cite{wang2022medclip} demonstrates superior performance in zero-shot prediction, supervised classification, and image-text retrieval tasks. MI-Zero \cite{lu2023visual}, a framework for zero-shot transfer in pathology, enables histopathological classification of images without additional labels by leveraging visual language pretraining. BioViL-T \cite{bannur2023learning} combines CNN-Transformer hybrid architectures for multi-image encoding with text models, improving semantic alignment between medical imaging and texts. KoBo \cite{chen2023knowledge} integrates clinical knowledge into contrastive pretraining. The CLIP-Driven Universal Model \cite{liu2023clipdriven} integrates text embeddings from CLIP with segmentation algorithms, delivering superior results in segmenting organs and detecting tumors in medical imaging.

\subsection{Mixture of Experts}

Recent explorations in LLMs have demonstrated the capability of MOEs \cite{jiang2024mixtral, zhou2022mixtureofexperts, fedus2022switch, du2022glam, zhang2023robust}. Puigcerver et al. introduce Soft MoE \cite{puigcerver2023sparse}, a fully-differentiable sparse Transformer with soft token-to-expert assignment, surpassing standard Transformers in visual recognition tasks. Zhou et al. developed Brainformer \cite{zhou2023brainformers}, integrating diverse layers including sparsely gated feed-forward layers, for efficient model scaling and faster convergence. Riquelme et al. introduce V-MoE \cite{NEURIPS2021_48237d9f}, a sparsely-gated Vision Transformer for image recognition, matching or surpassing state-of-the-art dense networks with reduced inference compute. Chowdhury et al. introduce pMoE \cite{chowdhury2023patchlevel}, a paradigm for CNN, enhancing efficiency through input patch division. Our MoME differs from these works in two fundamental ways:

\textit{Sparsity}: Prior methods use a sparse activation strategy to reduce inference time, engaging only the top-k experts. In contrast, MoME utilizes all experts during inference; each expert processes information relevant to its specialized function, ensuring comprehensive data integration for higher accuracy.
In mMIS, we prioritize precision over inference speed, aiming for maximum diagnostic accuracy.

\textit{Experts}: Former methods typically consist of several identical models. Our experts are distinct feature extractors at different \textbf{scales}, each tailored to capture specific scales of visual cues. This architecture promotes \textbf{diversity} among experts without specific assignments. 

\begin{figure*}[tb]
  \centering
  \includegraphics[width=0.9\textwidth]{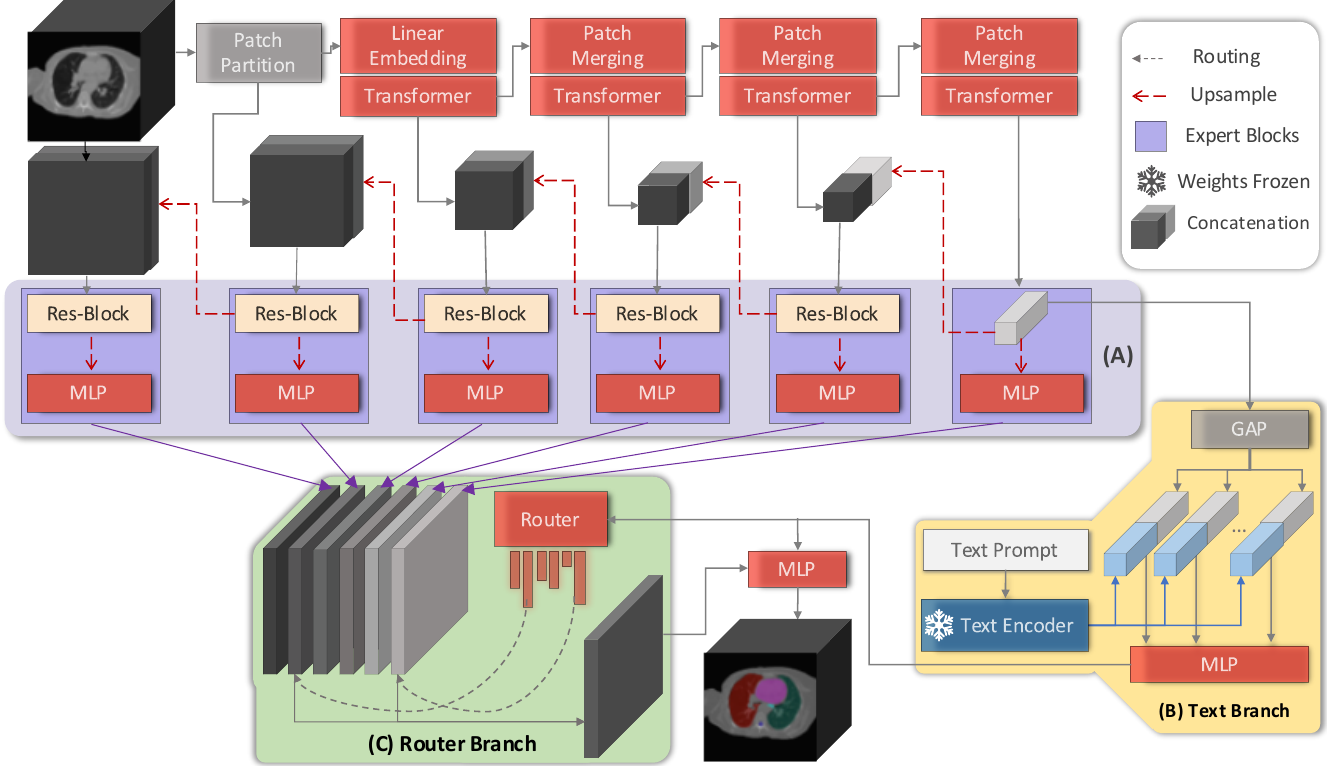}
  \caption{The architecture of MoME employs a text branch to generate CLIP embeddings for organs and tumors. Subsequently, a gating network is employed to intelligently direct the inputs from multiple experts, enabling effective integration of diverse visual features.
  }
  \label{fig:example}
\end{figure*}

\section{Methodology}
\label{sec:method}



Fig.\ref{fig:example} illustrates the architecture of MoME, which builds upon the foundation of the Universal model from \cite{liu2023clipdriven} by using a dynamic, text-guided expert aggregating mechanism, significantly enhancing segmentation precision.

The core innovation of MoME lies in its expert-driven architecture for MIS. 
Built on a Swin UNETR-like backbone with transformer blocks in the encoder and residual CNN blocks in the decoder, MoME uniquely treats each decoder layer as an expert specializing in scale-specific anatomical features (Fig \ref{fig:example}-A). Our novel routing mechanism (Fig \ref{fig:example}-C) then dynamically weights each decoder expert's contribution pixel-wise based on a combination of textual context and image feature from text branch (Fig \ref{fig:example}-B), enabling the model to adaptively focus on the most relevant expert for each anatomical structure, substantially improving segmentation performance across diverse medical imaging tasks.

\subsection{Problem Definition}
Following \cite{liu2023clipdriven}, we formulate the multi-dataset integration problem with $M$ datasets with a total of $N$ data points across $K$ distinct categories as $\mathcal{D} = \left\{\left(\boldsymbol{X}_i, \boldsymbol{Y}_i\right)\right\}_{i=1}^N$. Since these datasets originate from different sources with inconsistent labeling protocols, we face a partial labeling challenge where anatomical structures may be present but unannotated in certain scans. 

To address this challenge, we assign a rich semantic vector $\boldsymbol{w}_k$ to each class $k$ and design our model to solve $\mathcal{F}_\theta\left(\boldsymbol{X}_n, \boldsymbol{w}_k\right) = \boldsymbol{P}_n, n \in [1, N], k \in [1,K]$, across all samples and classes. We replace traditional one-hot encodings with CLIP embeddings for $\boldsymbol{w}_k$, which inherently capture anatomical relationships (e.g., between organs and their associated tumors), enabling MoME to perform robustly despite heterogeneous annotations across datasets.

\subsection{Visual Branch: Specialized Feature Extraction}
For CT scan preprocessing, we apply isotropic spacing and uniform intensity scaling to minimize the discrepancies across datasets. These prepared CT scans are input to the vision encoder.
We define $\boldsymbol{f}_l$ as the set of intermediate feature maps from each decoder layer, where $l \in \{1,2, ..., L\}$ indexes the layers. These features are then processed inside expert blocks.

\subsection{Expert Blocks}
Each expert in MoME builds upon one decoder layer, creating a multi-scale-expert architecture (Fig \ref{fig:example}-A). For each layer $l \in [1, L]$, $\boldsymbol{f}_l$ is upsampled and processed by a corresponding MLP to produce vision tokens $\boldsymbol{F} = \{\boldsymbol{F}_1, \boldsymbol{F}_2, ..., \boldsymbol{F}_L\}$ of uniform shape. Vision tokens are then utilized in the router branch.

\subsection{Text Branch: Textual Semantic Embedding}
Textual information provides crucial anatomical context that enhances segmentation accuracy. Our text branch leverages this insight by encoding rich semantic relationships between anatomical structures. For example, "hepatic vessel" and "liver" share substantial anatomical correlations that aren't captured by conventional approaches.

MoME's text branch (Fig \ref{fig:example}-B) processes each anatomical class through specialized medical prompts using a pre-trained CLIP text encoder. For class $k$, we generate embedding $\boldsymbol{w}_k$ using prompts such as "a computerized tomography of a [CLS]" where [CLS] represents the target structure. These embeddings are then combined with global image features $\boldsymbol{f}_L$ through a custom controller network:
$\boldsymbol{\theta}_k = \operatorname{MLP}(\boldsymbol{w}_k \oplus \boldsymbol{f}_L)$.
This integrated representation $\boldsymbol{\theta}_k$ is used in the dynamic segmentation head and guides our expert routing mechanism, enabling intelligent attention distribution across anatomical structures with varying visual characteristics.

\subsection{Router Branch: Token Selection and Fusion}
\textbf{Textual cues must be properly distributed to guide segmentation.} While hepatic vessel and liver are highly correlated \textbf{textually}, they differ \textbf{visually} in scales and anatomical structures, requiring distinct processing. MoE excels here by \textbf{dynamically routing} the textual cues to appropriate scales of feature maps. This automated routing allows MoME to adapt to the diverse requirements of each anatomical feature, ensuring effective translation of textual guidance into precise segmentation actions, as in our case, \textbf{pixel-wise} weights. 

At the heart of MoME's novelty is the gating network, or router $\mathcal{R}$ (Fig \ref{fig:example}-C). This branch integrates inputs from both the textual semantic embeddings $\boldsymbol{\theta}_k$ and the vision tokens $\boldsymbol{F} = \{\boldsymbol{F}_1, \boldsymbol{F}_2, ..., \boldsymbol{F}_L\}$ from expert blocks.
The router assesses the relevance and accuracy of each input in the context of the visual and textual tokens, dynamically allocating a gate-value $\boldsymbol{W}_l = \operatorname{MLP}(\boldsymbol{\theta}_k \oplus \boldsymbol{F}_l)$ to the most informative features.
To get the final output $G$, for each location $x$, the features are combined using $G(x) = \sum_{l=1}^{L} \frac{w_l \, (x)}{\sum_{j=1}^{L} w_j \, (x)} F_{l}(x)$.
This ensure that $G$ benefits from both deep semantic understanding from textual embeddings and the detailed visual information extracted by the vision branches and then processed inside expert blocks.

\subsection{Dynamic Segmentation Heads}
MoME employs specialized segmentation heads that adapt to each anatomical target based on the expert-aggregated representation $\boldsymbol{G}$. This approach allows the model to tailor its segmentation strategy according to the unique characteristics of each anatomical structure.

Our segmentation process applies three sequential $1 \times 1 \times 1$ convolutional operations to $\boldsymbol{G}$. The first two layers utilize 8 channels each, while the final layer produces a single-channel output for class $[\mathrm{CLS}]_k$. These operations are parameterized by the text-visual integration:
$\boldsymbol{P}_k = \operatorname{Sigmoid}\left(\left(\left(\boldsymbol{G} * \boldsymbol{\theta}_{k_1}\right) * \boldsymbol{\theta}_{k_2}\right) * \boldsymbol{\theta}_{k_3}\right)$.
Here, $\boldsymbol{\theta}_k = \left\{\boldsymbol{\theta}_{k_1}, \boldsymbol{\theta}_{k_2}, \boldsymbol{\theta}_{k_3}\right\}$ represents the parameters derived from our text module. This architecture produces predictions $\boldsymbol{P}_k \in \mathbb{R}^{1 \times D \times W \times H}$ for each anatomical class using a one-vs-all approach with sigmoid activation, effectively isolating each target structure while accounting for its unique anatomical characteristics.
\section{Experiments}
\label{sec:experiments}

\subsection{Implementation Details}
MoME is trained with the AdamW optimizer with a warm-up cosine scheduler over 50 epochs. We employ a batch size of 1 per GPU, and a patch dimension of $96 \times 96 \times 96$. The initial setup includes a learning rate of $1e^{-4}$, a momentum of 0.9, and a decay rate of $1e^{-5}$. This framework is developed using MONAI version 0.9.0, and trained on NVIDIA RTX 3090Ti graphic cards. 

Training data consists of an assembly of 10 public dataset (Pancreas-CT\cite{roth2015deeporgan}, LiTS\cite{bilic2019liver}, KiTS\cite{heller2020international}, AbdomenCT-1K\cite{ma2021abdomenct}, CT-ORG \cite{rister2020ct}, CHAOS\cite{valindria2018multi}, MSD\cite{antonelli2021medical}, BTCV\cite{landman2015miccai}, AMOS22 \cite{ji2022amos} and WORD \cite{luo2021word}), processed with a standardized labeling scheme.

\begin{table*}[t]
\centering
\caption{
Comparison of MoME-V and MoME-VL against other approaches, using 5-fold cross-validation on BTCV dataset. 
}
\scriptsize 
\begin{tabular}{p{0.18\linewidth}P{0.03\linewidth}P{0.03\linewidth}P{0.03\linewidth}P{0.03\linewidth}P{0.03\linewidth}P{0.03\linewidth}P{0.03\linewidth}P{0.03\linewidth}P{0.03\linewidth}P{0.03\linewidth}P{0.03\linewidth}P{0.03\linewidth}P{0.03\linewidth}}
\hline
Methods & Spl  
& RKid & LKid 
& Gall  & Eso 
& Liv & Sto 
& Aor & IVC 
& Veins   & Pan 
& AG & Avg. \\ 
\hline

RandPatch~\cite{tang2021high}     
& 95.82 & 88.52                        
& 90.14 & 68.31                        
& 75.01 & 96.48                        
& 82.93 & 88.96   
& 82.49 & 73.54
& 75.48 & 66.09
& 80.76
\\ 
TransBTS~\cite{wang2021transbts}     
& 94.59 & 89.23                     
& 90.47 & 68.50                       
& 75.59 & 96.14                       
& 83.72 & 88.85  
& 82.28 & 74.25
& 75.12 & 66.74 
& 80.94
\\ 
nnFormer~\cite{zhou2021nnformer}     
& 94.51 & 88.49                     
& 93.39 & 65.51                        
& 74.49 & 96.10                      
& 83.83 & 88.91
& 80.58 & 75.94
& 77.71 & {68.19}
& 81.22
\\
UNETR~\cite{hatamizadeh2022unetr}
& 94.91 & 92.10                        
& 93.12 & 76.98                        
& 74.01 & 96.17                        
& 79.98 & 89.74   
& 81.20 & 75.05
& 80.12 & 62.60
& 81.43
\\
nnU-Net~\cite{isensee2021nnu}     
& \textbf{95.92} & 88.28                        
& 92.62 & 66.58                        
& 75.71 & 96.49                        
& 86.05 & 88.33   
& 82.72 & {78.31}
& 79.17 & 67.99
& 82.01
\\ \hline
SwinUNETR~\cite{tang2022self}    
& 95.44 & {93.38}                 
& 93.40 & 77.12                    
& 74.14 & 96.39                        
& 80.12 & 90.02   
& 82.93 & 75.08
& 81.02 & 64.98
& 82.06
\\

MoME-V & 95.65 & {93.66} & {94.01} & {78.25} & {75.64} & {96.53} & {83.52} & {90.36} & {84.11} & 75.28 & {81.86} & {65.65} & {84.54}
\\ \hline
Universal Model \cite{liu2023clipdriven} & 95.82 & {94.28} & {94.11} & {79.52} & {76.55} & \textbf{97.05} & {92.59} & {91.63} & {86.00} & 77.54 & {83.17} & {70.52} & {86.13}
\\

MoME-VL & 95.87 & \textbf{94.45} & \textbf{94.34} & \textbf{80.63} & \textbf{78.52} & {97.03} & \textbf{92.99} & \textbf{91.97} & \textbf{86.82} & \textbf{78.93} & \textbf{83.65} & \textbf{71.32} & \textbf{87.21}
\\ \hline
\end{tabular}%
\label{tab:btcv_benchmark}
\end{table*}

\begin{table*}[t]
\caption
{\textbf{Tumor detection performance.} LiTS~\cite{bilic2019liver}, KiTS~\cite{heller2019kits19}, and MSD Pancreas~\cite{antonelli2021medical} contain tumors in the liver, kidney, and pancreas, respectively, and are used to compute tumor detection sensitivity (Sen.). To assess specificity (Spec.), we use CHAOS~\cite{valindria2018multi} and Pancreas-CT~\cite{roth2015deeporgan}, which contain no tumors in the corresponding organs. The harmonic mean (Harm.) is reported to reflect the balance between sensitivity and specificity. 
}
\centering
\scriptsize
\begin{tabular}{p{0.17\linewidth}|P{0.04\linewidth}P{0.04\linewidth}P{0.04\linewidth}|P{0.04\linewidth}P{0.04\linewidth}P{0.04\linewidth}|P{0.04\linewidth}P{0.04\linewidth}P{0.04\linewidth}}
\hline
\multirow{2}{*}{Methods} & \multicolumn{3}{c|}{Liver Tumor} & \multicolumn{3}{c|}{Kidney Tumor} &\multicolumn{3}{c}{Pancreatic Tumor} \\
& Sen. & Spec. & Harm.
& Sen. & Spec. & Harm.
& Sen. & Spec. & Harm.
\\ \hline
nnU-Net~\cite{isensee2021nnu} & \textbf{94.44} & 75.00 & 83.60 & 96.88 & 85.00 & 90.55 & 95.18 & 88.75 & 91.85
\\
UNet++~\cite{zhou2019unet++}& \textbf{94.44} & 80.00 & 86.62 &N/A&N/A&N/A&N/A & N/A&N/A
\\
UNETR~\cite{hatamizadeh2022unetr} & 86.11 & \textbf{95.00} & 90.34 & 93.75 & \textbf{95.00} & \textbf{94.37} & 90.36 & 81.25 & 85.56
\\ \hline
Swin UNETR~\cite{tang2022self} & 91.67 & 85.00 & 88.21 & \textbf{97.91} & 70.00 & 81.63 & 97.59 & 87.50 & 92.26
\\

MoME-V & 90.37 & 90.00 & 90.18 & 97.04 & 75.00 & 84.61 & \textbf{97.64} & 87.50 & 92.29
\\ \hline
Universal Model \cite{liu2023clipdriven} & 88.89 & \textbf{95.00} & 91.84 & 91.67 & \textbf{95.00} & 93.31 & 93.98 & 91.25 & 92.59
\\ 
MoME-VL & 89.15 & \textbf{95.00} & \textbf{91.98} & 92.01 & \textbf{95.00} & 93.48 & 93.66 & \textbf{92.50} & \textbf{93.08}
\\ \hline
\end{tabular}%
\label{tab:high_specificity}
\end{table*}

\begin{table*}[h]
\caption{\textbf{\textit{Generalizability:}} Performance evaluation on \textbf{external datasets}, conducted without further fine-tuning or domain-specific adjustments. \textit{mDSC*} represents the mean Dice score across the first seven organs. MoME-VL demonstrates superior robustness on CT images.}
\centering
\scriptsize
\begin{tabular}{p{0.18\linewidth}P{0.03\linewidth}P{0.03\linewidth}P{0.03\linewidth}P{0.03\linewidth}P{0.03\linewidth}P{0.03\linewidth}P{0.03\linewidth}P{0.03\linewidth}P{0.03\linewidth}P{0.04\linewidth}P{0.03\linewidth}}
\hline
3D-IRCADb & Spl & RKid & LKid
& Gall & Liv & Sto 
& Pan & RLun & LLun & mDSC* & mDSC
\\ 
\hline
SegResNet~\cite{siddiquee2021redundancy} & 94.08 & 80.01 & 91.60 & 69.59 & 95.62 & {89.53} & 79.19 & N/A & N/A & 85.66 & N/A
\\
nnFormer~\cite{zhou2021nnformer} & 93.75 & 88.20 & 90.11 & 62.22 & 94.93 & 87.93 & 78.90  & N/A & N/A & 85.14 & N/A
\\
UNesT~\cite{yu2022unest} & 94.02 & 84.90 & {94.95} & 68.58 & 95.10 & 89.28 & 79.94 & N/A & N/A & 86.68 & N/A
\\
TransBTS~\cite{wang2021transbts} & 91.33 & 76.22 & 88.87 & 62.50 & 94.42 & 85.87 & 63.90 & N/A & N/A & 80.44 & N/A
\\
TransUNet~\cite{chen2021transunet} & 94.09 & 82.07 & 89.92 & 63.07 & 95.55 & 89.12 & 79.53 & N/A & N/A & 84.76 & N/A
\\
UNETR~\cite{hatamizadeh2022unetr} & 92.23 & 91.28 & 94.19 & 56.20 & 94.25 & 86.73 & 72.56 & 91.56 & 93.31 & 83.92 & 85.81
\\
\hline
Swin UNETR~\cite{tang2022self} & 93.51 & 66.34 & 90.63 & 61.05 & 94.73 & 87.37 & 73.77 & 93.72 & 92.17 & 81.05 & 83.69
\\
MoME-V & 93.95 & 76.24 & 91.66 & 62.45 & 95.34 & 87.57 & 74.62 & 95.32 & 94.88 & 84.54 & 85.78
\\
\hline
Universal Model~\cite{liu2023clipdriven} & {95.76} & {94.99} & 94.42 & {88.79} & {97.03} & 89.36 & {80.99} & {97.71} & {96.72} & {91.62} & {92.86}
\\
MoME-VL & \textbf{96.42} & \textbf{95.78} & \textbf{95.23} & \textbf{89.22} & \textbf{97.07} & \textbf{90.25} & \textbf{81.45} & \textbf{98.00} & \textbf{97.52} & \textbf{92.20} & \textbf{93.44}
\\
\hline
& & & & & & & & & & &
\end{tabular}%

\begin{tabular}{p{0.18\linewidth}P{0.08\linewidth}P{0.05\linewidth}P{0.07\linewidth}P{0.09\linewidth}P{0.08\linewidth}P{0.04\linewidth}P{0.05\linewidth}}

Thorax-85 & Esophagus & Trachea & Vertebrae & Right Lung & Left Lung & Heart & mDSC
\\ 
\hline
nnU-Net~\cite{isensee2021nnu} & \textbf{81.18} & 89.32 & \textbf{91.21} & 97.74 & 97.68 & 92.66 & 91.63 
\\
TransUNet~\cite{chen2021transunet} & 78.27 &  91.45 & 88.36 & 97.84 & 97.63 & \textbf{94.74} & 91.38
\\
UNETR~\cite{hatamizadeh2022unetr} & 79.37 & 91.09 & 88.02 & 97.23 & 97.11 & 94.23 & 91.18
\\
MedSAM~\cite{ma2023segment} & 80.25 & 91.78 & 89.12 & 97.81 & 97.55 & 94.16 & 91.78
\\
Med-SA~\cite{wu2023medical} & 77.91 & 91.19 & 90.12 & 97.70 & \textbf{97.85} & 94.23 & 91.50
\\
\hline
Swin UNETR~\cite{tang2022self} & 79.64 & 91.28 & 88.89 & 97.69 & 97.64 & 94.66 & 91.63
\\
MoME-V & 80.18 & 91.64 & 89.05 & 97.86 & 97.32 & 94.71 & 91.79
\\
\hline
Universal Model~\cite{liu2023clipdriven} & 79.89 & 91.87 & 88.62 & 97.32 & 97.35 & 94.27 & 91.55
\\
MoME-VL & 80.25 & \textbf{92.03} & {89.23} & \textbf{97.93} & 97.82 & {94.23} & \textbf{91.92}
\\
\hline
\end{tabular}%
\label{tab:generalizability}
\end{table*}

\begin{table}[h]
\caption%
{\textbf{\textit{Transferability:} Fine-tuning performance.} Fine-tuning MoME-VL yields considerably superior results compared to training from scratch on the TotalSegmentator. Additionally, by employing image segmentation as a proxy task, MoME-VL is capable of deriving a more refined visual representation than other models pre-trained in the medical field.}

\centering
\scriptsize
\begin{tabular}{p{0.29\linewidth}|P{0.07\linewidth}P{0.07\linewidth}P{0.07\linewidth}P{0.07\linewidth}}
\hline
Method & vertebrae & cardiac & muscles & organs 
\\ \hline
Scratch & 81.06&84.47&88.83&86.42
\\
MedicalNet~\cite{chen2019med3d}&82.28&87.40&91.36&86.90
\\
Models Gen.~\cite{zhou2019models}&85.12&86.51&89.96&85.78
\\
Swin UNETR~\cite{tang2022self}&86.23&87.91&92.39&88.56
\\
UniMiSS~\cite{xie2022unimiss}&85.12&88.96&92.86&88.51
\\
Universal Model~\cite{liu2023clipdriven} & 86.49 & {89.57} & {94.43} & {88.95} 
\\
\hline
MoME-VL & \textbf{86.92} & \textbf{90.41} & \textbf{94.98} & \textbf{89.62} 
\\
\hline
\end{tabular}%
\label{tab:transfer_learning}
\end{table}

\begin{table*}[t]
\centering
\caption{Ablation study on Top-K router choice: The comparison is based on a 5-fold cross-validation performed on the BTCV dataset, using the Dice score as the evaluation criterion.}

\scriptsize 
\begin{tabular}{p{0.01\linewidth}|P{0.035\linewidth}P{0.035\linewidth}P{0.035\linewidth}P{0.035\linewidth}P{0.035\linewidth}P{0.035\linewidth}P{0.035\linewidth}P{0.035\linewidth}P{0.035\linewidth}P{0.035\linewidth}P{0.035\linewidth}P{0.035\linewidth}P{0.035\linewidth}}
\hline
K & Spl  
& RKid & LKid 
& Gall  & Eso 
& Liv & Sto 
& Aor & IVC 
& Veins   & Pan 
& AG & Avg. \\ 
\hline

1   
& 85.47&91.00&86.91&74.72&73.39&90.76&86.58&85.15&80.81&70.52&75.21&60.52&80.09
\\
2 & 89.15&93.48&89.81&76.15&75.57&93.68&87.72&88.04&81.74&74.81&78.42&65.37&82.83
\\ 
3 & 93.32&93.75&92.71&80.07&77.88&95.94&91.79&90.24&85.55&76.75&82.72&69.23&85.83
\\ 
4 & 95.36&93.90&93.61&80.17&77.91&96.93&92.44&91.65&86.00&78.49&83.18&70.77&86.70
\\ 
5 & 95.74&94.25&94.26&80.30&78.11&96.72&92.75&91.70&86.80&78.66&83.37&70.92&86.97
\\ 
6 & \textbf{95.87} & \textbf{94.45} & \textbf{94.34} & \textbf{80.63} & \textbf{78.52} & \textbf{97.03} & \textbf{92.99} & \textbf{91.97} & \textbf{86.82} & \textbf{78.93} & \textbf{83.65} & \textbf{71.32} & \textbf{87.21}
\\ 
\hline
\end{tabular}%
\label{tab:router_choice}
\end{table*}

\subsection{Quantitative Results}

Our 5-fold cross-validation on the BTCV dataset (Table \ref{tab:btcv_benchmark}) yields insightful observations; The transition from Swin UNETR to MoME-V (solely uses visual information) showcases marked improvements in segmentation accuracy for specific organs.
Notably, liver (Liv) Dice scores rose from 96.39\% to 96.53\%, and aorta (Aor) from 80.12\% to 83.52\%.
These results highlight the effectiveness of incorporating MOE into Swin UNETR, which allows for enhanced adaptation and specialization in the segmentation of these organs. The average Dice score saw an improvement from 82.06\% with Swin UNETR to 84.54\% with MoME-V, emphasizing the added benefit of MOE to the base architecture, thereby improving overall segmentation performance across various organs.

Transitioning from the Universal Model to MoME-VL (integrates both visual and textual data) illustrates the benefit of integrating textual embeddings with MOE. This combination significantly improves segmentation accuracy, especially for anatomically complex organs such as the pancreas (Pan), with accuracy improving from 83.17\% to 83.65\%, and the adrenal gland (AG) from 70.52\% to 71.32\%. MoME-VL achieved the highest average Dice score of 87.21\%, surpassing the Universal Model's 86.13\%. This improvement highlights the powerful synergy of leveraging textual context for enhanced understanding and MOE for increased adaptability, leading to superior segmentation accuracy across all evaluated organs.

\subsubsection{Tumor Detection Performance}
AI systems may yield a high number of false positives for tumor-free CT scans.
 Consequently, we assess patient-level Sensitivity and Specificity in identifying three tumor types, and report their harmonic mean to showcase the equilibrium between detection capabilities. Tumor-free CT scans are sourced from CHAOS and Pancreas-CT, both pathologically confirmed to be tumor-free. According to Table \ref{tab:high_specificity}, MoME-VL yields harmonic means of 91.98\%, 93.48\%, and 93.08\% across the three tumor types, demonstrating high sensitivity for tumor detection while maintaining minimal false positives to achieve a balanced performance. Compared to models trained on single datasets, MoME-VL’s lower false positive rate demonstrates the benefits of integrating diverse datasets, offering richer and more plentiful positive and negative examples for improved accuracy and reliability.


\subsubsection{Generalizability to External Datasets} Effective medical AI systems must generalize across diverse datasets from various hospitals, rather than being optimized for a single dataset. As shown in Table \ref{tab:generalizability}, MoME-VL benefits from training on a significantly broader range of CT scans, resulting in superior generalizability compared to models limited to specific datasets. We evaluate MoME-VL on two fully held-out datasets—3D-IRCADb \cite{soler20103d} (public) and Thorax-85 (private)—neither of which was used during training. MoME-VL performs robustly across both (Table \ref{tab:generalizability}), achieving a 93.44\% mean Dice score (mDSC) on 3D-IRCADb without further adjustments. Its adaptability is further demonstrated on Thorax-85, with scores of 80.25\% for the Esophagus and 92.03\% for the Trachea. These results reflect the strength of MoME-VL’s advanced design, which integrates Swin UNETR, textual embeddings, and a Mixture of Experts (MoE) to enhance both anatomical understanding and cross-dataset generalization in medical imaging.

\subsubsection{Transferability to Unseen Organs} 
MoME-VL demonstrates its efficacy as a robust pre-training model for segmentation tasks. Pre-trained on a composite dataset and fine-tuned on multiple targets, MoME-VL achieves top Dice scores of 86.92\%, 90.41\%, 94.98\%, and 89.62\% across four downstream tasks in the TotalSegmentator dataset (refer to Table \ref{tab:transfer_learning}).

\begin{figure*}[tbh!]
  \centering
  \includegraphics[width=\textwidth]{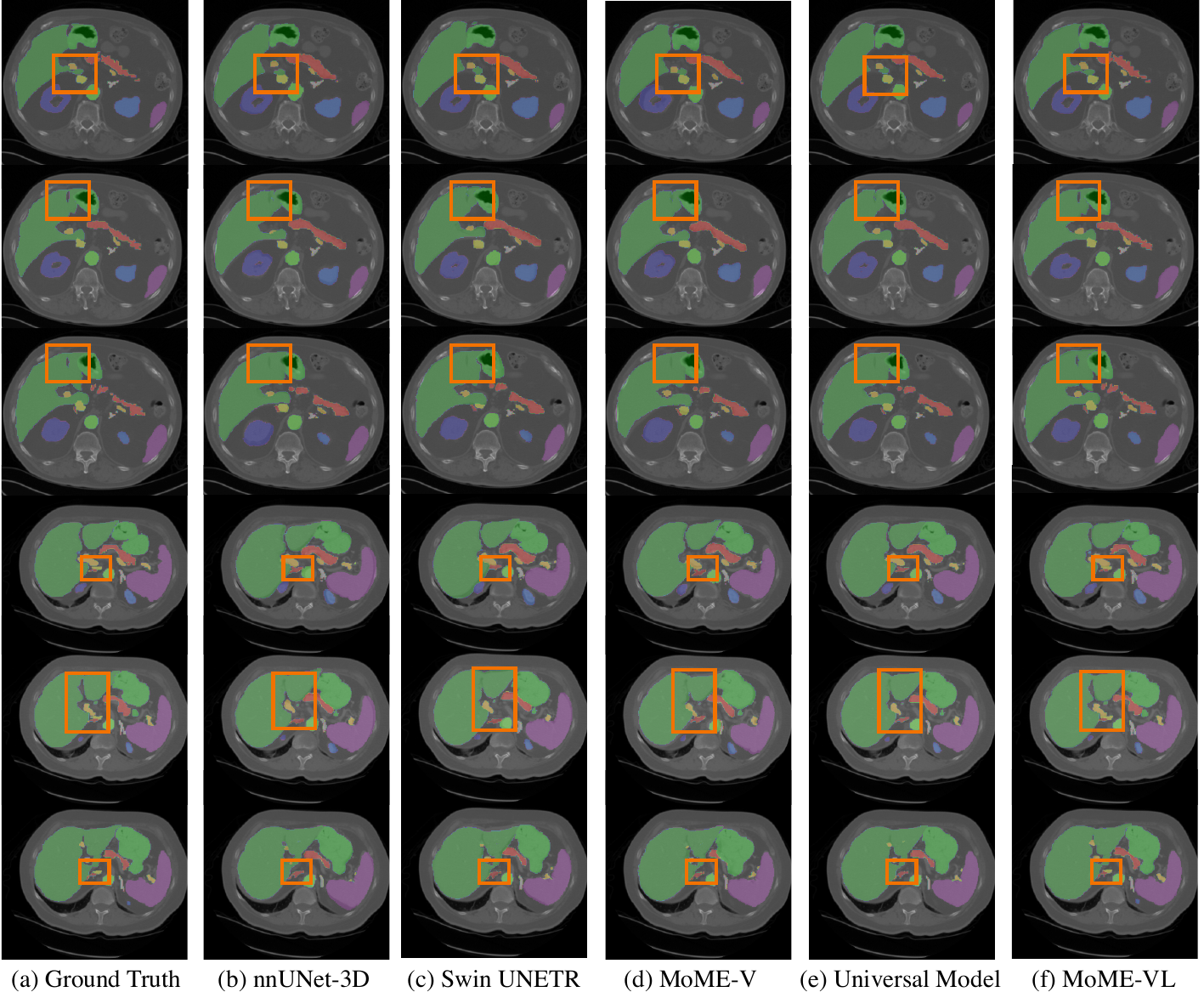}
  \caption{Qualitative analysis of various methods on the BTCV dataset: Column (a) displays the ground truth. Columns (b), (c), and (e) display results from earlier techniques, while (d) and (f) show MoME's performance. Orange rectangles highlight our model's effectiveness.}
  \label{fig:qualitative}
\end{figure*}

\subsubsection{The Impact of Expert Quantity in Router Configuration} 
An ablation study was conducted to assess the impact of the number of experts in the router on overall performance. 
Table \ref{tab:router_choice} shows that increasing the number of experts consistently improves Dice scores across all organs.
This suggests that the segmentation model benefits from diversified expertise, with each expert potentially specializing in different aspects of the task, thereby improving the model's ability to accurately delineate anatomical structures.
Increasing the number of experts provides the model with broader perspectives for segmentation, enhancing the model's adaptability and precision in handling the complex variability inherent in medical imaging data. 

\subsection{Qualitative Analysis}
Fig \ref{fig:qualitative} showcases the qualitative results from different approaches on the BTCV dataset, with MoME-VL displaying exceptional performance over competing methods. The areas highlighted by orange rectangles emphasize the superior capabilities of our model. As illustrated in the final column, it delivers uniform and accurate predictions.

\section{Conclusion and Future Directions}
\label{sec:conclusion}
In this study, we introduce MoME: Mixture of Visual Language Medical Experts for segmenting medical images. By integrating the Mixture of Experts (MoE) approach with textual embeddings into the segmentation framework, our model excels at learning from datasets with incomplete labeling and delivers superior performance. Crucially, we demonstrate that combining MoE with visual language embeddings can foster a deeper and more significant understanding of the anatomical relationships between different organs, addressing the fundamental challenges of anatomical diversity and contextual complexity in medical imaging.

MoME utilizes multi-scale experts, each built atop a single decoder layer. We believe our MoME framework provides a strong foundation for advancing medical image segmentation. Future work could extend this approach by treating entire models (e.g., Swin UNETRs) as individual experts to further increase model diversity.


\printbibliography

@String(ICLR = {Int. Conf. Learn. Represent.})

@String(AAAI = {AAAI})

@String(ICLR  = {ICLR})

@article{antonelli2021medical,
  title={The medical segmentation decathlon},
  author={Antonelli, Michela and Reinke, Annika and Bakas, Spyridon and Farahani, Keyvan and Landman, Bennett A and Litjens, Geert and Menze, Bjoern and Ronneberger, Olaf and Summers, Ronald M and van Ginneken, Bram and others},
  journal={arXiv preprint arXiv:2106.05735},
  year={2021}
}

@article{bilic2019liver,
  title={The liver tumor segmentation benchmark (lits)},
  author={Bilic, Patrick and Christ, Patrick Ferdinand and Vorontsov, Eugene and Chlebus, Grzegorz and Chen, Hao and Dou, Qi and Fu, Chi-Wing and Han, Xiao and Heng, Pheng-Ann and Hesser, J{\"u}rgen and others},
  journal={arXiv preprint arXiv:1901.04056},
  year={2019}
}

@article{chen2021transunet,
  title={Transunet: Transformers make strong encoders for medical image segmentation},
  author={Chen, Jieneng and Lu, Yongyi and Yu, Qihang and Luo, Xiangde and Adeli, Ehsan and Wang, Yan and Lu, Le and Yuille, Alan L and Zhou, Yuyin},
  journal={arXiv preprint arXiv:2102.04306},
  year={2021}
}

@article{chen2019med3d,
  title={Med3d: Transfer learning for 3d medical image analysis},
  author={Chen, Sihong and Ma, Kai and Zheng, Yefeng},
  journal={arXiv preprint arXiv:1904.00625},
  year={2019}
}

@inproceedings{hatamizadeh2022unetr,
  title={UNETR: Transformers for 3d medical image segmentation},
  author={Hatamizadeh, Ali and Tang, Yucheng and Nath, Vishwesh and Yang, Dong and Myronenko, Andriy and Landman, Bennett and Roth, Holger R and Xu, Daguang},
  booktitle={Proceedings of the IEEE/CVF Winter Conference on Applications of Computer Vision},
  pages={574--584},
  year={2022}
}

@misc{heller2020international,
  title={An international challenge to use artificial intelligence to define the state-of-the-art in kidney and kidney tumor segmentation in ct imaging.},
  author={Heller, Nicholas and McSweeney, Sean and Peterson, Matthew Thomas and Peterson, Sarah and Rickman, Jack and Stai, Bethany and Tejpaul, Resha and Oestreich, Makinna and Blake, Paul and Rosenberg, Joel and others},
  year={2020}
}

@article{heller2019kits19,
  title={The kits19 challenge data: 300 kidney tumor cases with clinical context, ct semantic segmentations, and surgical outcomes},
  author={Heller, Nicholas and Sathianathen, Niranjan and Kalapara, Arveen and Walczak, Edward and Moore, Keenan and Kaluzniak, Heather and Rosenberg, Joel and Blake, Paul and Rengel, Zachary and Oestreich, Makinna and others},
  journal={arXiv preprint arXiv:1904.00445},
  year={2019}
}

@article{isensee2021nnu,
  title={nnU-Net: a self-configuring method for deep learning-based biomedical image segmentation},
  author={Isensee, Fabian and Jaeger, Paul F and Kohl, Simon AA and Petersen, Jens and Maier-Hein, Klaus H},
  journal={Nature Methods},
  volume={18},
  number={2},
  pages={203--211},
  year={2021},
  publisher={Nature Publishing Group}
}

@article{ji2022amos,
  title={Amos: A large-scale abdominal multi-organ benchmark for versatile medical image segmentation},
  author={Ji, Yuanfeng and Bai, Haotian and Yang, Jie and Ge, Chongjian and Zhu, Ye and Zhang, Ruimao and Li, Zhen and Zhang, Lingyan and Ma, Wanling and Wan, Xiang and others},
  journal={Neural Information Processing Systems (NeurIPS)},
  year={2022}
}

@inproceedings{landman2015miccai,
  title={Miccai multi-atlas labeling beyond the cranial vault--workshop and challenge},
  author={Landman, Bennett and Xu, Zhoubing and Igelsias, J and Styner, Martin and Langerak, T and Klein, Arno},
  booktitle={Proc. MICCAI Multi-Atlas Labeling Beyond Cranial Vault—Workshop Challenge},
  volume={5},
  pages={12},
  year={2015}
}

@article{luo2021word,
  title={Word: Revisiting organs segmentation in the whole abdominal region},
  author={Luo, Xiangde and Liao, Wenjun and Xiao, Jianghong and Song, Tao and Zhang, Xiaofan and Li, Kang and Wang, Guotai and Zhang, Shaoting},
  journal={arXiv preprint arXiv:2111.02403},
  year={2021}
}

@article{ma2021abdomenct,
  title={Abdomenct-1k: Is abdominal organ segmentation a solved problem},
  author={Ma, Jun and Zhang, Yao and Gu, Song and Zhu, Cheng and Ge, Cheng and Zhang, Yichi and An, Xingle and Wang, Congcong and Wang, Qiyuan and Liu, Xin and others},
  journal={IEEE Transactions on Pattern Analysis and Machine Intelligence},
  year={2021}
}

@article{rister2020ct,
  title={Ct-org, a new dataset for multiple organ segmentation in computed tomography},
  author={Rister, Blaine and Yi, Darvin and Shivakumar, Kaushik and Nobashi, Tomomi and Rubin, Daniel L},
  journal={Scientific Data},
  volume={7},
  number={1},
  pages={1--9},
  year={2020},
  publisher={Nature Publishing Group}
}

@inproceedings{roth2015deeporgan,
  title={Deeporgan: Multi-level deep convolutional networks for automated pancreas segmentation},
  author={Roth, Holger R and Lu, Le and Farag, Amal and Shin, Hoo-Chang and Liu, Jiamin and Turkbey, Evrim B and Summers, Ronald M},
  booktitle={International conference on medical image computing and computer-assisted intervention},
  pages={556--564},
  organization={Springer},
  year={2015}
}

@article{siddiquee2021redundancy,
  title={Redundancy reduction in semantic segmentation of 3d brain tumor mris},
  author={Siddiquee, Md Mahfuzur Rahman and Myronenko, Andriy},
  journal={arXiv preprint arXiv:2111.00742},
  year={2021}
}

@techreport{soler20103d,
  title={3d image reconstruction for comparison of algorithm database: A patient specific anatomical and medical image database},
  author={Soler, L and Hostettler, A and Agnus, V and Charnoz, A and Fasquel, J and Moreau, J and Osswald, A and Bouhadjar, M and Marescaux, J},
  year={2010},
  institution={IRCAD, Strasbourg, France}
}

@article{tang2021high,
  title={High-resolution 3d abdominal segmentation with random patch network fusion},
  author={Tang, Yucheng and Gao, Riqiang and Lee, Ho Hin and Han, Shizhong and Chen, Yunqiang and Gao, Dashan and Nath, Vishwesh and Bermudez, Camilo and Savona, Michael R and Abramson, Richard G and others},
  journal={Medical image analysis},
  volume={69},
  pages={101894},
  year={2021},
  publisher={Elsevier}
}

@inproceedings{tang2022self,
  title={Self-supervised pre-training of swin transformers for 3d medical image analysis},
  author={Tang, Yucheng and Yang, Dong and Li, Wenqi and Roth, Holger R and Landman, Bennett and Xu, Daguang and Nath, Vishwesh and Hatamizadeh, Ali},
  booktitle={Proceedings of the IEEE/CVF Conference on Computer Vision and Pattern Recognition},
  pages={20730--20740},
  year={2022}
}

@inproceedings{valindria2018multi,
  title={Multi-modal learning from unpaired images: Application to multi-organ segmentation in ct and mri},
  author={Valindria, Vanya V and Pawlowski, Nick and Rajchl, Martin and Lavdas, Ioannis and Aboagye, Eric O and Rockall, Andrea G and Rueckert, Daniel and Glocker, Ben},
  booktitle={2018 IEEE winter conference on applications of computer vision (WACV)},
  pages={547--556},
  organization={IEEE},
  year={2018}
}

@inproceedings{wang2021transbts,
  title={Transbts: Multimodal brain tumor segmentation using transformer},
  author={Wang, Wenxuan and Chen, Chen and Ding, Meng and Yu, Hong and Zha, Sen and Li, Jiangyun},
  booktitle={International Conference on Medical Image Computing and Computer-Assisted Intervention},
  pages={109--119},
  organization={Springer},
  year={2021}
}

@inproceedings{xie2022unimiss,
  title={Unimiss: Universal medical self-supervised learning via breaking dimensionality barrier},
  author={Xie, Yutong and Zhang, Jianpeng and Xia, Yong and Wu, Qi},
  booktitle={European Conference on Computer Vision},
  pages={558--575},
  organization={Springer},
  year={2022}
}

@article{yu2022unest,
  title={Unest: Local spatial representation learning with hierarchical transformer for efficient medical segmentation},
  author={Yu, Xin and Yang, Qi and Zhou, Yinchi and Cai, Leon Y and Gao, Riqiang and Lee, Ho Hin and Li, Thomas and Bao, Shunxing and Xu, Zhoubing and Lasko, Thomas A and others},
  journal={arXiv preprint arXiv:2209.14378},
  year={2022}
}

@article{zhou2021nnformer,
  title={nnformer: Interleaved transformer for volumetric segmentation},
  author={Zhou, Hong-Yu and Guo, Jiansen and Zhang, Yinghao and Yu, Lequan and Wang, Liansheng and Yu, Yizhou},
  journal={arXiv preprint arXiv:2109.03201},
  year={2021}
}

@article{zhou2019unet++,
  title={Unet++: Redesigning skip connections to exploit multiscale features in image segmentation},
  author={Zhou, Zongwei and Siddiquee, Md Mahfuzur Rahman and Tajbakhsh, Nima and Liang, Jianming},
  journal={IEEE transactions on medical imaging},
  volume={39},
  number={6},
  pages={1856--1867},
  year={2019},
  publisher={IEEE}
}

@inproceedings{zhou2019models,
  title={Models genesis: Generic autodidactic models for 3d medical image analysis},
  author={Zhou, Zongwei and Sodha, Vatsal and Siddiquee, Md Mahfuzur Rahman and Feng, Ruibin and Tajbakhsh, Nima and Gotway, Michael B and Liang, Jianming},
  booktitle={International conference on medical image computing and computer-assisted intervention},
  pages={384--393},
  organization={Springer},
  year={2019}
}

@misc{fedus2022switch,
      title={Switch Transformers: Scaling to Trillion Parameter Models with Simple and Efficient Sparsity}, 
      author={William Fedus and Barret Zoph and Noam Shazeer},
      year={2022},
      eprint={2101.03961},
      archivePrefix={arXiv},
      primaryClass={cs.LG}
}

@misc{du2022glam,
      title={GLaM: Efficient Scaling of Language Models with Mixture-of-Experts}, 
      author={Nan Du and Yanping Huang and Andrew M. Dai and Simon Tong and Dmitry Lepikhin and Yuanzhong Xu and Maxim Krikun and Yanqi Zhou and Adams Wei Yu and Orhan Firat and Barret Zoph and Liam Fedus and Maarten Bosma and Zongwei Zhou and Tao Wang and Yu Emma Wang and Kellie Webster and Marie Pellat and Kevin Robinson and Kathleen Meier-Hellstern and Toju Duke and Lucas Dixon and Kun Zhang and Quoc V Le and Yonghui Wu and Zhifeng Chen and Claire Cui},
      year={2022},
      eprint={2112.06905},
      archivePrefix={arXiv},
      primaryClass={cs.CL}
}

@inproceedings{NEURIPS2021_48237d9f,
 author = {Riquelme, Carlos and Puigcerver, Joan and Mustafa, Basil and Neumann, Maxim and Jenatton, Rodolphe and Susano Pinto, Andr\'{e} and Keysers, Daniel and Houlsby, Neil},
 booktitle = {Advances in Neural Information Processing Systems},
 editor = {M. Ranzato and A. Beygelzimer and Y. Dauphin and P.S. Liang and J. Wortman Vaughan},
 pages = {8583--8595},
 publisher = {Curran Associates, Inc.},
 title = {Scaling Vision with Sparse Mixture of Experts},
 volume = {34},
 year = {2021}
}

@misc{zhou2022mixtureofexperts,
      title={Mixture-of-Experts with Expert Choice Routing}, 
      author={Yanqi Zhou and Tao Lei and Hanxiao Liu and Nan Du and Yanping Huang and Vincent Zhao and Andrew Dai and Zhifeng Chen and Quoc Le and James Laudon},
      year={2022},
      eprint={2202.09368},
      archivePrefix={arXiv},
      primaryClass={cs.LG}
}

@misc{zhou2023brainformers,
      title={Brainformers: Trading Simplicity for Efficiency}, 
      author={Yanqi Zhou and Nan Du and Yanping Huang and Daiyi Peng and Chang Lan and Da Huang and Siamak Shakeri and David So and Andrew Dai and Yifeng Lu and Zhifeng Chen and Quoc Le and Claire Cui and James Laundon and Jeff Dean},
      year={2023},
      eprint={2306.00008},
      archivePrefix={arXiv},
      primaryClass={cs.LG}
}

@misc{puigcerver2023sparse,
      title={From Sparse to Soft Mixtures of Experts}, 
      author={Joan Puigcerver and Carlos Riquelme and Basil Mustafa and Neil Houlsby},
      year={2023},
      eprint={2308.00951},
      archivePrefix={arXiv},
      primaryClass={cs.LG}
}

@misc{xue2024openmoe,
      title={OpenMoE: An Early Effort on Open Mixture-of-Experts Language Models}, 
      author={Fuzhao Xue and Zian Zheng and Yao Fu and Jinjie Ni and Zangwei Zheng and Wangchunshu Zhou and Yang You},
      year={2024},
      eprint={2402.01739},
      archivePrefix={arXiv},
      primaryClass={cs.CL}
}

@misc{chowdhury2023patchlevel,
      title={Patch-level Routing in Mixture-of-Experts is Provably Sample-efficient for Convolutional Neural Networks}, 
      author={Mohammed Nowaz Rabbani Chowdhury and Shuai Zhang and Meng Wang and Sijia Liu and Pin-Yu Chen},
      year={2023},
      eprint={2306.04073},
      archivePrefix={arXiv},
      primaryClass={cs.LG}
}

@misc{zhang2023robust,
      title={Robust Mixture-of-Expert Training for Convolutional Neural Networks}, 
      author={Yihua Zhang and Ruisi Cai and Tianlong Chen and Guanhua Zhang and Huan Zhang and Pin-Yu Chen and Shiyu Chang and Zhangyang Wang and Sijia Liu},
      year={2023},
      eprint={2308.10110},
      archivePrefix={arXiv},
      primaryClass={cs.CV}
}

@misc{jiang2024mixtral,
      title={Mixtral of Experts}, 
      author={Albert Q. Jiang and Alexandre Sablayrolles and Antoine Roux and Arthur Mensch and Blanche Savary and Chris Bamford and Devendra Singh Chaplot and Diego de las Casas and Emma Bou Hanna and Florian Bressand and Gianna Lengyel and Guillaume Bour and Guillaume Lample and Lélio Renard Lavaud and Lucile Saulnier and Marie-Anne Lachaux and Pierre Stock and Sandeep Subramanian and Sophia Yang and Szymon Antoniak and Teven Le Scao and Théophile Gervet and Thibaut Lavril and Thomas Wang and Timothée Lacroix and William El Sayed},
      year={2024},
      eprint={2401.04088},
      archivePrefix={arXiv},
      primaryClass={cs.LG}
}

@article{ExpertlevelTiu,
	Author = {Tiu, Ekin and Talius, Ellie and Patel, Pujan and Langlotz, Curtis P. and Ng, Andrew Y. and Rajpurkar, Pranav},
	Da = {2022/12/01},
	Id = {Tiu2022},
	Isbn = {2157-846X},
	Journal = {Nature Biomedical Engineering},
	Number = {12},
	Pages = {1399--1406},
	Title = {Expert-level detection of pathologies from unannotated chest X-ray images via self-supervised learning},
	Ty = {JOUR},
	Volume = {6},
	Year = {2022},
	}

@misc{wang2022medclip,
      title={MedCLIP: Contrastive Learning from Unpaired Medical Images and Text}, 
      author={Zifeng Wang and Zhenbang Wu and Dinesh Agarwal and Jimeng Sun},
      year={2022},
      eprint={2210.10163},
      archivePrefix={arXiv},
      primaryClass={cs.CV}
}

@misc{lu2023visual,
      title={Visual Language Pretrained Multiple Instance Zero-Shot Transfer for Histopathology Images}, 
      author={Ming Y. Lu and Bowen Chen and Andrew Zhang and Drew F. K. Williamson and Richard J. Chen and Tong Ding and Long Phi Le and Yung-Sung Chuang and Faisal Mahmood},
      year={2023},
      eprint={2306.07831},
      archivePrefix={arXiv},
      primaryClass={cs.CV}
}

@misc{bannur2023learning,
      title={Learning to Exploit Temporal Structure for Biomedical Vision-Language Processing}, 
      author={Shruthi Bannur and Stephanie Hyland and Qianchu Liu and Fernando Pérez-García and Maximilian Ilse and Daniel C. Castro and Benedikt Boecking and Harshita Sharma and Kenza Bouzid and Anja Thieme and Anton Schwaighofer and Maria Wetscherek and Matthew P. Lungren and Aditya Nori and Javier Alvarez-Valle and Ozan Oktay},
      year={2023},
      eprint={2301.04558},
      archivePrefix={arXiv},
      primaryClass={cs.CV}
}

@misc{chen2023knowledge,
      title={Knowledge Boosting: Rethinking Medical Contrastive Vision-Language Pre-Training}, 
      author={Xiaofei Chen and Yuting He and Cheng Xue and Rongjun Ge and Shuo Li and Guanyu Yang},
      year={2023},
      eprint={2307.07246},
      archivePrefix={arXiv},
      primaryClass={cs.CV}
}

@misc{liu2023clipdriven,
      title={CLIP-Driven Universal Model for Organ Segmentation and Tumor Detection}, 
      author={Jie Liu and Yixiao Zhang and Jie-Neng Chen and Junfei Xiao and Yongyi Lu and Bennett A. Landman and Yixuan Yuan and Alan Yuille and Yucheng Tang and Zongwei Zhou},
      year={2023},
      eprint={2301.00785},
      archivePrefix={arXiv},
      primaryClass={eess.IV}
}

@article{Yan_Pei_2022, title={Clinical-BERT: Vision-Language Pre-training for Radiograph Diagnosis and Reports Generation}, volume={36}, abstractNote={In this paper, we propose a vision-language pre-training model, Clinical-BERT, for the medical domain, and devise three domain-specific tasks: Clinical Diagnosis (CD), Masked MeSH Modeling (MMM), Image-MeSH Matching (IMM), together with one general pre-training task: Masked Language Modeling (MLM), to pre-train the model. The CD task helps the model to learn medical domain knowledge by predicting disease from radiographs. Medical Subject Headings (MeSH) words are important semantic components in radiograph reports, and the MMM task helps the model focus on the prediction of MeSH words. The IMM task helps the model learn the alignment of MeSH words with radiographs by matching scores obtained by a two-level sparse attention: region sparse attention and word sparse attention. Region sparse attention generates corresponding visual features for each word, and word sparse attention enhances the contribution of images-MeSH matching to the matching scores. To the best of our knowledge, this is the first attempt to learn domain knowledge during pre-training for the medical domain. We evaluate the pre-training model on Radiograph Diagnosis and Reports Generation tasks across four challenging datasets: MIMIC-CXR, IU X-Ray, COV-CTR, and NIH, and achieve state-of-the-art results for all the tasks, which demonstrates the effectiveness of our pre-training model.}, number={3}, journal={Proceedings of the AAAI Conference on Artificial Intelligence}, author={Yan, Bin and Pei, Mingtao}, year={2022}, month={Jun.}, pages={2982-2990} }

@misc{wu2023medical,
      title={Medical SAM Adapter: Adapting Segment Anything Model for Medical Image Segmentation}, 
      author={Junde Wu and Wei Ji and Yuanpei Liu and Huazhu Fu and Min Xu and Yanwu Xu and Yueming Jin},
      year={2023},
      eprint={2304.12620},
      archivePrefix={arXiv},
      primaryClass={cs.CV}
}

@misc{ma2023segment,
      title={Segment Anything in Medical Images}, 
      author={Jun Ma and Yuting He and Feifei Li and Lin Han and Chenyu You and Bo Wang},
      year={2023},
      eprint={2304.12306},
      archivePrefix={arXiv},
      primaryClass={eess.IV}
}

@misc{vorontsov2024virchow,
      title={Virchow: A Million-Slide Digital Pathology Foundation Model}, 
      author={Eugene Vorontsov and Alican Bozkurt and Adam Casson and George Shaikovski and Michal Zelechowski and Siqi Liu and Kristen Severson and Eric Zimmermann and James Hall and Neil Tenenholtz and Nicolo Fusi and Philippe Mathieu and Alexander van Eck and Donghun Lee and Julian Viret and Eric Robert and Yi Kan Wang and Jeremy D. Kunz and Matthew C. H. Lee and Jan Bernhard and Ran A. Godrich and Gerard Oakley and Ewan Millar and Matthew Hanna and Juan Retamero and William A. Moye and Razik Yousfi and Christopher Kanan and David Klimstra and Brandon Rothrock and Thomas J. Fuchs},
      year={2024},
      eprint={2309.07778},
      archivePrefix={arXiv},
      primaryClass={eess.IV}
}

@inproceedings{noroozi2016unsupervised,
  title={Unsupervised learning of visual representations by solving jigsaw puzzles},
  author={Noroozi, Mehdi and Favaro, Paolo},
  booktitle={European Conference on Computer Vision},
  pages={69--84},
  year={2016},
  organization={Springer}
}

@inproceedings{komodakis2018unsupervised,
  title={Unsupervised representation learning by predicting image rotations},
  author={Komodakis, Nikos and Gidaris, Spyros},
  booktitle={ICLR},
  year={2018}
}

@inproceedings{doersch2015unsupervised,
  title={Unsupervised visual representation learning by context prediction},
  author={Doersch, Carl and Gupta, Abhinav and Efros, Alexei A},
  booktitle={Proceedings of the IEEE International Conference on Computer Vision},
  pages={1422--1430},
  year={2015}
}

@inproceedings{vaswani2017attention,
  title={Attention is all you need},
  author={Vaswani, Ashish and Shazeer, Noam and Parmar, Niki and Uszkoreit, Jakob and Jones, Llion and Gomez, Aidan N and Kaiser, {\L}ukasz and Polosukhin, Illia},
  booktitle={Advances in neural information processing systems},
  pages={5998--6008},
  year={2017}
}

@article{gibson2018automatic,
  title={Automatic multi-organ segmentation on abdominal CT with dense v-networks},
  author={Gibson, Eli and Giganti, Francesco and Hu, Yipeng and Bonmati, Ester and Bandula, Steve and Gurusamy, Kurinchi and Davidson, Brian and Pereira, Stephen P and Clarkson, Matthew J and Barratt, Dean C},
  journal={IEEE transactions on medical imaging},
  volume={37},
  number={8},
  pages={1822--1834},
  year={2018},
  publisher={IEEE}
}

@inproceedings{ronneberger2015unet,
  title={U-net: Convolutional networks for biomedical image segmentation},
  author={Ronneberger, Olaf and Fischer, Philipp and Brox, Thomas},
  booktitle={International Conference on Medical image computing and computer-assisted intervention},
  pages={234--241},
  year={2015},
  organization={Springer}
}

\end{document}